\documentclass[10pt, twocolumn]{article}

\usepackage{cite}
\usepackage[pdftex]{graphicx}
\usepackage{amsmath}
\usepackage[table, dvipsnames]{xcolor}
\usepackage{algorithmic}
\usepackage[caption=false,font=footnotesize]{subfig}
\usepackage{stfloats}

\usepackage{fullpage}
\usepackage{authblk}
\usepackage{pdflscape}
\usepackage{afterpage}

\usepackage[T1]{fontenc}
\usepackage[]{lmodern}
\usepackage[utf8]{inputenc}
\usepackage[english]{babel}

\usepackage{multirow}
\usepackage{multicol}
\usepackage{hyperref}
\hypersetup{
  colorlinks   = true, 
  urlcolor     = blue, 
  linkcolor    = blue, 
  citecolor   = red 
}

\def\C#1{\multicolumn{1}{c|}{#1}}

\newcommand{\ie}{\textit{i.e.}}

\definecolor{LightCyan}{rgb}{0.88,1,1}

\begin{document}

\title{Comparing two deep learning sequence-based models for protein-protein interaction prediction}

\author[1,2]{Florian~Richoux}
\author[1]{Charlène~Servantie}
\author[3]{Cynthia~Borès}
\author[3]{Stéphane~Téletchéa}
\affil[1]{LS2N, Université de Nantes, France}
\affil[ ]{\it florian.richoux@univ-nantes.fr}
\affil[2]{JFLI, CNRS, National Institute of Informatics, Japan}
\affil[3]{UFIP, Université de Nantes, France}
\affil[ ]{\it stephane.teletchea@univ-nantes.fr}

\maketitle

\begin{abstract}
Biological data are extremely diverse, complex but also quite sparse. The recent developments in deep learning methods are offering new possibilities for the analysis of complex data. However, it is easy to be get a deep learning model that seems to have good results but is in fact either overfitting the training data or the validation data. In particular, the fact to overfit the validation data, called "information leak", is almost never treated in papers proposing deep learning models to predict protein-protein interactions (PPI). In this work, we compare two carefully designed deep learning models and show pitfalls to avoid while predicting PPIs through machine learning methods. Our best model predicts accurately more than 78\% of human PPI, in very strict conditions both for training and testing. The methodology we propose here allow us to have strong confidences about the ability of a model to scale up on larger datasets. This would allow sharper models when larger datasets would be available, rather than current models prone to information leaks. Our solid methodological foundations shall be applicable to more organisms and whole proteome networks predictions.
\end{abstract}

\section{Introduction}

Machine learning methods are extensively used in biology to complement experimental data, otherwise hard and costly to acquire~\cite{Chen2015}\cite{Topol2014}. Machine learning methods were successfully applied to a wide range of biological purposes such as large-scale ecosystem reconstruction~\cite{TARA2015}, whole-cell modelling~\cite{ABASY2016}, pathway analysis and integration~\cite{Pathways2005}, and the prediction of drug side-effects~\cite{PharmacologySEA2007}. The recent progress in deep learning have unraveled new possibilities for reportedly difficult predictions in biology, as was recently demonstrated for protein structure prediction by AlphaFold~\cite{Service2018}. One of the most interesting advantage of deep learning compare to classical machine learning methods is the automation of feature extractions: previously, the training process was based on features one had to furnish to a model, like it is the case with Support Vector Machine for instance~\cite{Shen2007}. However, it is not always clear what good features one must furnish to a model, and recent deep learning results show that machines can be better than humans to extract and identify usefull features hidden among a large amont of raw data.

Out of the successful domains of applications indicated above, there is still a significant bottleneck in cell biology modelling coming from the incomplete description of protein-protein interactions (PPI), \ie, the interaction of two proteins at the molecular level, thus preventing a comprehensive description of living cells~\cite{PPI2016}. Up to now, PPI data was available from various sources built upon experimental analyses or predictions, but a comprehensive database was recently set up to assemble properly this knowledge by a combination of validated experimental data and profile-kernel Support Vector Machine analysis~\cite{Tran2018}.

It is important to rely on robust experimental data to build a reliable model, this is why we set up our dataset from the expert annotation of protein-protein interactions in humans as available in uniprot~\cite{Uniprot2017}. This restricted but reliable dataset was further divided into a training set, into a hold-out validation set, and into a hold-out test set. We have set up two deep learning models and compare them to have a better view of pitfalls to avoid while studying the prediction of protein-protein interactions through machine learning methods. One of our models shows robustness against both overfitting and information leak, which is usually poorly treated in other papers.

\section{Methods}

To allow a complete reproducibility of this work, our datasets, source
code,   experimental    setup   and    results   are    available   at
\href{https://gitlab.univ-nantes.fr/richoux-f/DeepPPI/tree/v1.tcbb}{https://gitlab.univ-nantes.fr/richoux-f/DeepPPI/tree/v1.tcbb}.

\subsection{Datasets}

The Uniprot  web site was queried  on June 18th, 2018  to retrieve all
human    sequences   where    a   protein-protein    interaction   was
indicated. This  led to an  initial list of  validated protein-protein
interactions  for  more than 47,000  couples  of proteins. An in-house
python script was developed using biopython~\cite{Biopython2000}\cite{Biopython2009} to produce  a  negative dataset  from a  random
picking of  proteins, where it  was first checked that  an interaction
was   not  present. This led to a preliminary set of 96,106  couples   of
proteins where each protein is represented by its chain of amino acid residues only.

From  this  set,  we  extracted  proteins composed  of  at  most  1166
amino acids.   This   represents  68,334  couples  of   proteins.   We
randomized couples in this set and divided it into three distinct sets
exactly composed of  50\% of positive and negative  samples each. Randomization  before making  these three sets  allow to
not have a series of the same protein within one set.

These three sets are: the training set (52,606 couples - 26,303
positives and  26,303 negatives),  the hold-out validation  set (6,574
couples - 3287 positives and 3287 negatives) and the hold-out test set
(6,574  couples  - 3287  positives  and  3287 negatives).   Having  an
hold-out validation set is  better than making cross-validation, since
we have a completely dedicated set for the validation of samples that never   appear  into  the  training   process.  However,  a
validation  set only  is  not sufficient:  during the  hyper-parameters
optimization process, one tries  to find hyper-parameters that
fit  the validation  set.  Even  without being  directly  part of  the
training process, a  model can still overfit the  validation set. This
behavior is known  as ``information  leak''. To prevent  this form  of indirect
overfitting, a hold-out test set is use to perform final test once the
model and its hyper-parameters are fixed. Again, samples from the test
set  never  appear   neither in  the  training  nor in the hyper-parameters optimization processes.  

We call  these training/validation/test  sets our {\it  regular} sets.
However, the regular test set has a possible flaw: even if  none of its
couples of  proteins appear  neither into the training  nor the
validation set, most of its proteins  individually appear into the training
and/or the  validation set. To prevent  a sort of overfitting  where a
model could learn for instance that "protein X never interacts", we
also  made three  other sets,  named  {\it strict}  sets, obtained  as
follow: we isolate couples composed of a
protein  that appears  at  most  twice in  the  whole dataset.   These
extracted couples constitute our test set (460 samples - 230 positives
and 230 negatives). The remaining samples  are split into two parts to
give our  training set (57,722  samples - 28,861 positives  and 28,861
negatives) and  validation set  (6,412 samples  - 3,206  positives and
3,206 negatives).

Finally, we augmented  both our regular and strict sets  by adding the
mirror copy of  each couple: if a set contains  the couple (protein A,
protein B),  we add it the mirror  couple (protein B,  protein A),
with of  course the same  label "interaction / no  interaction". Since
our original  dataset already contains  a few of mirror  couples, this
transformation    does    not    exactly double   our
sets. Table~\ref{tab:datasets} indicates our new datasets size. Naturally,
these  sets remain  composed  of  50\%   positive  samples and  of 50\%
negative samples.  Modulo mirror  copies, all  couples of  proteins in
each set  are unique:  we did  not do  bootstrapping to  increase the
number of our samples.

\begin{table}[ht]
  \centering
  {
    \begin{tabular}{||c|c|c||c|c|c||}
      \hline
      \multicolumn{3}{||c||}{Regular} & \multicolumn{3}{c||}{Strict}\\
      \hline
      train & val. & test & train & val. & test\\
      85,104 & 12,822 & 12,806 & 91,036 & 12,506 & 720\\
      \hline
    \end{tabular}
  }
  \caption{Number of protein couples in our final datasets}
  \label{tab:datasets}
\end{table}

\subsection{Models}

\begin{figure}
  \centering
  \includegraphics[width=0.9\linewidth]{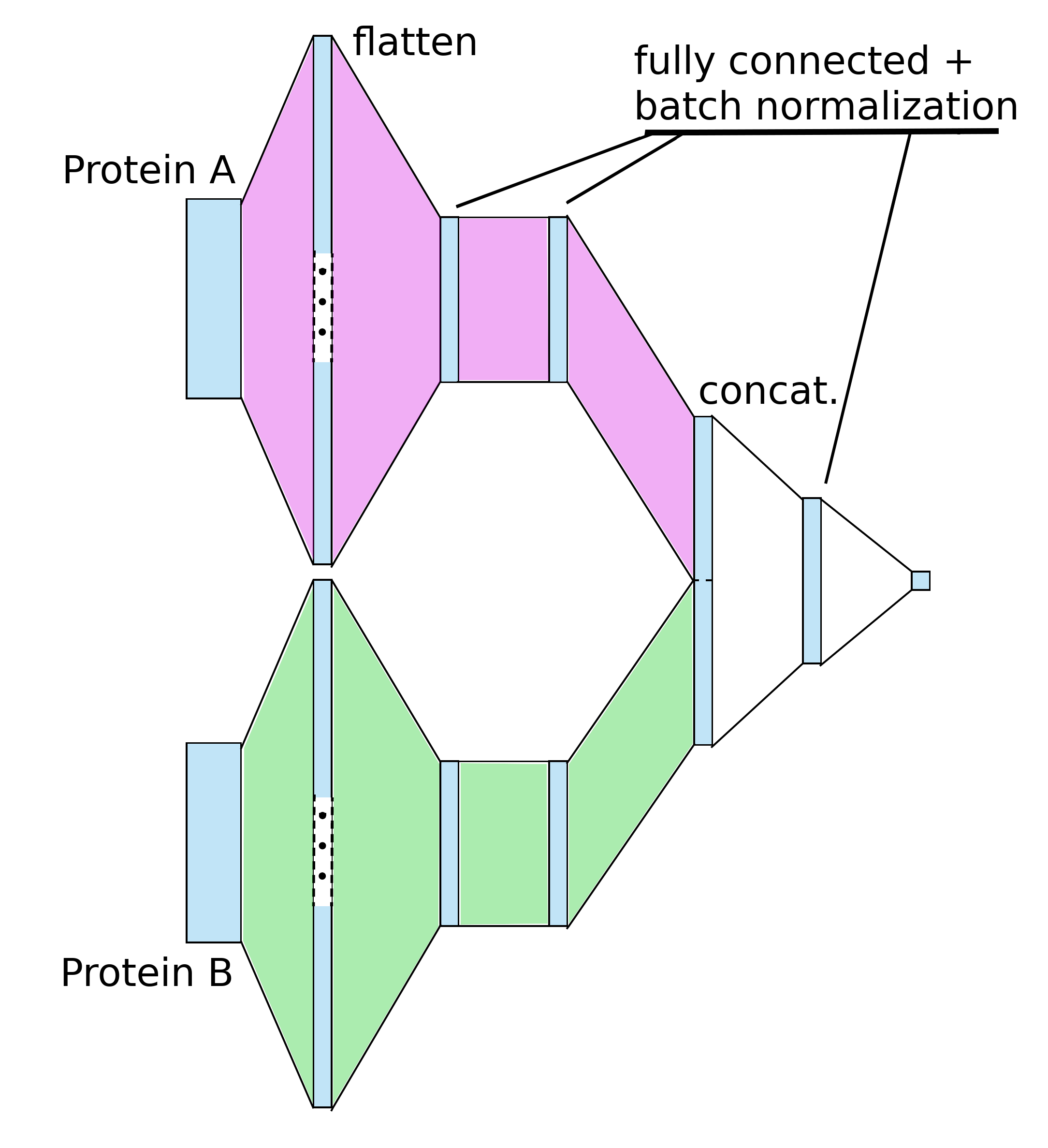}
  \caption{Our fully connected model.  Parts in cyan are layers of neurons. Weights of the feature extractions part of the network are independent for Protein A (pink part) and Protein B (green part).}
  \label{fig:fc_flat}
\end{figure}

In this work, we propose to compare  two different neural network architectures:
a fully  connected model  and a recurrent  model. Our  fully connected
model  is  illustrated  by Figure~\ref{fig:fc_flat}.     This  model  has  a total  of
1,121,481 parameters,  \ie, the network  is composed of  1,121,481 weights
learned   during  the   training   process.    Our  recurrent   model,
illustrated by~Figure~\ref{fig:lstm}, is more than a 100
times smaller  with 10,090 parameters.   Despite seeming to  have worst
results than our fully connected model (but at first glance only, like
discussed   in   Section~\ref{sec:results}),   this  model   is   more
interesting since ``simpler'', in the  sense it has significantly less
parameters and then  would be more scalable when we  will increase its
size if  we can train it  on a larger training  set.  Moreover, having
more parameters than training samples is often leading to overfitting.
Favoring small  models is a  way to get a  more robust network  that will
generalize   well   on   new   data.    As   shown   in
Table~\ref{tab:datasets}, our training  sets contain about 90,000
samples.

\subsubsection{Input representation}

We recall that our inputs are restricted  to the chain of amino acid residues of two
given  proteins,  the goal  being  to  predict  if such  proteins  can
interact each  other or  not. We  model a protein input as
a sequence  of vectors of 24 Boolean values, one vector for each amino acid of the protein sequence. These  vectors are one-hot encoding  amino acids: they are  true  only at  the  index  characterizing  the amino  acid.   For
instance,  let's assume that the index  for  Alanine  (A)  is  4 (an arbitrary value), then  then  vector
representing it  is a 24-long  vector filled by zeros  (or false)
except at index 4  where the value is one  (or true). In addition
of  the  20 standard  proteinogenic  amino  acids, our  datasets  also
contain  selenocysteine (U),  a placeholder  for either  asparagine or
aspartic acid (B),  another one for either glutamic  acid or glutamine
(Z) and a placeholder for unknown acid (X). Since we consider proteins with a maximal sequence length of 1166 amino acids, and that most of proteins are shorter than that, our inputs are padded, allowing us to have the same input shape whatever the protein. The padding here is simply adding vectors of zeros until each protein is represented by 1166 vectors. Since these vectors have a length of 24, all proteins are represented by a matrix of shape (1166, 24).

\subsubsection{Our fully connected model architecture}

We explain here the choices we made to build our  models, layer by layer,
starting  by our  fully connected  model. Despite  being huge compared to our dataset size,  with a
total of 1,121,481 parameters, this  model is conceptually the simplest:
features extraction is  done for both proteins  sequence separately by
flattening inputs and feeding them to two fully connected layers. Then
we concatenate these layers and process to the classification with two
fully connected layers, as  shown Figure~\ref{fig:fc_flat}. Each fully
connected layers in the model  is followed  by a  batch normalization  layer.  This
allow a regulation  of our model, preventing  overfitting and speeding
up training time. Although theoretical reasons why batch normalization
help  are still  unclear~\cite{Shibani2018},  it  is well-known  that
batch normalization  and dropout  layers are  not giving  good results
when applied together~\cite{LiBatch2018}, this is the reason
why our  models do not  contain any dropout  layers in favor  of batch
normalization layers. Hyper-parameters of  this model are presented in
Table~\ref{tab:fc_model}.   All fully  connected layers  have 20  units
followed  by a  classical  ReLU activation  function, except  for the
final layer having one unit with a sigmoid activation for doing binary
classification (given proteins can either interact with each other or not).

\begin{table*}[ht]
  \centering
  {
    \begin{tabular}{llrc}
      \hline
      Layer & Hyper-parameters & Parameters & Output shape\\
      \hline
      \hline
      Input & Sequence length=1166 & - & (1166, 24)\\
      \hline
      Flatten & - & 0 & (27984)\\
      \hline
      Fully connected & Units=20, Activation=relu & 559,700 & (20)\\
      \hline
      Batch normalization & - & 80 & (20)\\
      \hline
      Fully connected & Units=20, Activation=relu & 420 & (20)\\
      \hline
      Batch normalization & - & 80 & (20)\\
      \hline
      Concatenation & - & 0& (40)\\
      \hline
      Fully connected & Units=20, Activation=relu & 820 & (20)\\
      \hline
      Batch normalization & - & 80 & (20)\\
      \hline
      Fully connected & Units=1, Activation=sigmoid & 21 & (1)\\
      \hline
    \end{tabular}
  }
  \caption{Our fully connected model hyper-parameters,
    parameters and tensor sizes.}
  \label{tab:fc_model}
\end{table*}

\begin{figure*}
  \centering
  \includegraphics[width=\linewidth]{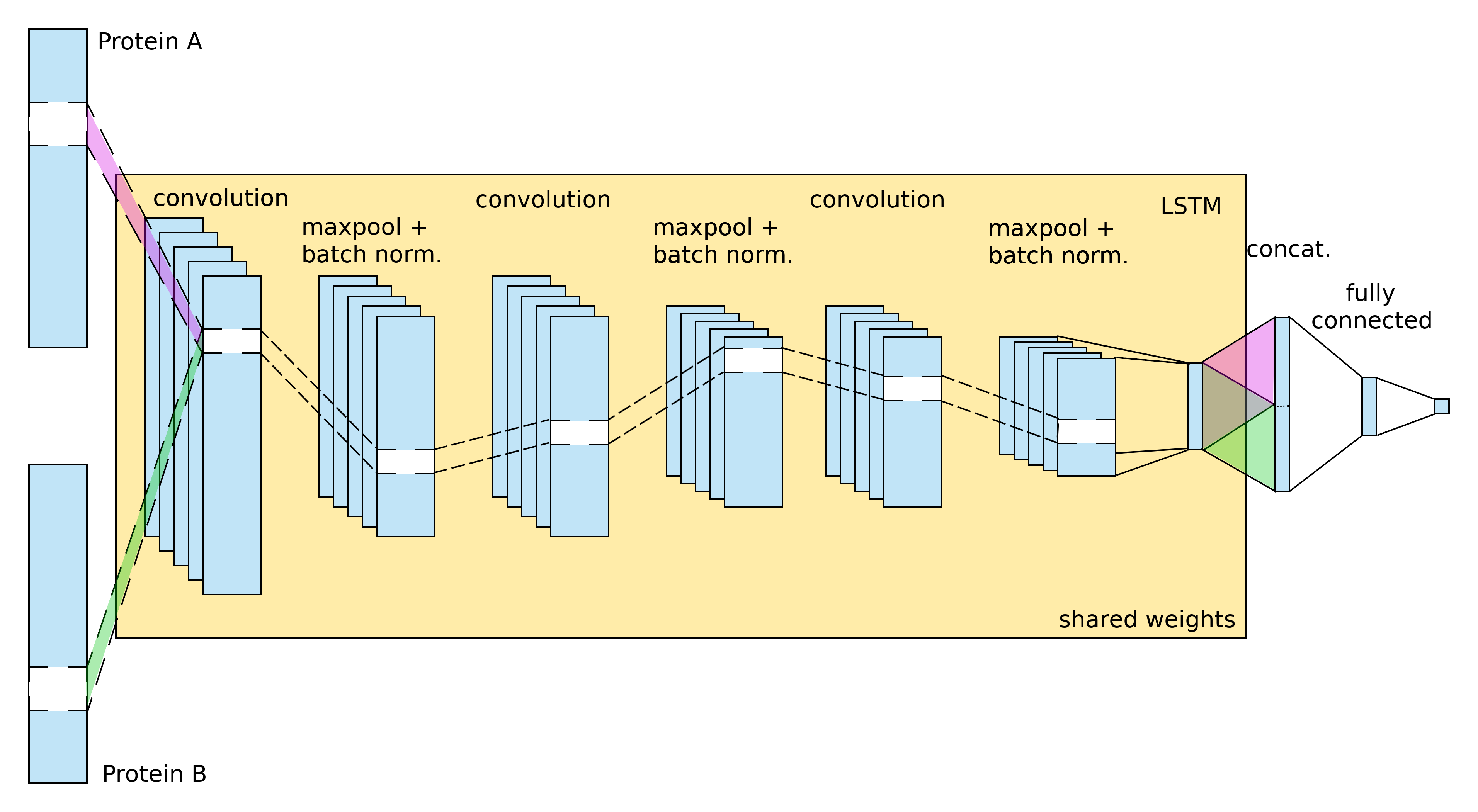}
  \caption{Our recurrent  model. Parts in cyan are layers of neurons. Layers in the orange frame
    are sharing weights for both proteins.}
  \label{fig:lstm}
\end{figure*}

\subsubsection{Our recurrent model architecture}

Our  recurrent layers  is  smaller but  not  as straight-forward,  see
Figure~\ref{fig:lstm}. Features extraction is done for each protein by
a  sequence  of  three  one-dimension  convolution  +  pooling  +  batch
normalization closed by  a recurrent Long Short-Term  Memory layer, or
LSTM.   Parameters  of these  extraction  features  layers are  shared
between our two  inputs, dividing by two the number  of parameters for
this part of  the network and reinforcing weights  learning, since the
same  layers  must  deal  with the chain of amino acid residues  of  both  proteins.
Indeed, there are no reasons to think that features for the first input
must be different from features for the second one. The reader shall notice tha we tried to
also share parameters  for extraction features in  our fully connected
model, but this  led to slightly worst  results.  Here, one-dimension
convolutions     have     a     kernel      size     of     20,     see
Table~\ref{tab:lstm_model}. It  means the  first convolution  layer is
looking at a window of 20 amino acids and outputs some value regarding
this window, then  repeats this operation by shifting  its window from
just  one amino  acid (stride  of 1).   Follows  a pooling  layer
taking the convolution layer output and squashing each series of three
values in  a row by selecting  the maximal value.  The  combination of
these convolution and  pooling layers allow both to  extract the local
features among successive amino acids  and to significantly reduce the
input  shape of  next layers,  making the  network smaller,  faster to
train  and more  prone to be generalized.  Like  for our  fully connected
model,  batch  normalization layers  are  here  to both  regulate  and
accelerate the training process. A ReLU activation function is applied
after  each convolution  layer.  Features  extraction finishes  with a
small LSTM layer with 32 units.  Like all recurrent layers, LSTM allows
to extract  spatial and temporal  features from sequences.   Since our
inputs are  chains of amino acid residues, there is  no notion of time,  but we
clearly  have  a  spatial  dimension  to  take  into  account:  it  is
reasonable to consider that amino  acids arrangement is significant for
having two proteins interacting each  other. The series of convolution
and pooling layers  extract some local spatial features,  but LSTM can
extract   global   spatial  features   on   the   whole  sequence   by
``remembering'' and  then considering  previously treated  elements. A
hyperbolic  tangent activation  function is  performed after  the LSTM
layer, which is  also a classical choice for such  layer. Once feature
extraction is  done, we concatenate  data and  give them to  two fully
connected layers for performing  the classification.  This part of
the network,  usually referred  to be  the ``head''  of the  model, is
identical to the head of our fully connected model, modulo the number
of units  of the first fully  connected layer (here, 25),

\begin{table*}[ht]
  \centering
  \small
  {
    \begin{tabular}{llrc}
      \hline
      Layer & Hyper-parameters & Parameters & Output shape\\
      \hline
      \hline
      Input & Sequence length=1166 & - & (1166, 24)\\
      \hline
      \rowcolor{LightCyan}
      {\bf Convolution 1D} & {\bf Filters=5, Kernel size=20, Stride=1,
                             Activation=relu} & {\bf 2,405} & {\bf (1147, 5)}\\
      \hline
      \rowcolor{LightCyan}
      {\bf MaxPooling 1D} & {\bf Pool size=3}& {\bf 0}& {\bf (382, 5)}\\
      \hline
      \rowcolor{LightCyan}
      {\bf Batch normalization} & {\bf -} & {\bf 20} & {\bf (382, 5)}\\
      \hline
      \rowcolor{LightCyan}
      {\bf Convolution 1D} & {\bf Filters=5, Kernel size=20, Stride=1,
                             Activation=relu} & {\bf 505} & {\bf (363, 5)}\\
      \hline
      \rowcolor{LightCyan}
      {\bf MaxPooling 1D} & {\bf Pool size=3} & {\bf 0} & {\bf (121, 5)}\\
      \hline
      \rowcolor{LightCyan}
      {\bf Batch normalization} & {\bf -} & {\bf 20} & {\bf (121, 5)}\\
      \hline
      \rowcolor{LightCyan}
      {\bf Convolution 1D} & {\bf Filters=5, Kernel size=20, Stride=1,
                             Activation=relu} & {\bf 505} & {\bf (102, 5)}\\
      \hline
      \rowcolor{LightCyan}
      {\bf MaxPooling 1D} & {\bf Pool size=3} & {\bf 0} & {\bf (34, 5)}\\
      \hline
      \rowcolor{LightCyan}
      {\bf Batch normalization} & {\bf -} & {\bf 20} & {\bf (34, 5)}\\
      \hline
      \rowcolor{LightCyan}
      {\bf LSTM} & {\bf Units=32, Activation=tanh} & {\bf 4,864} & {\bf
                                                                  (32)}\\
      \hline
      Concatenation & - & 0& (64)\\
      \hline
      Fully connected & Units=25, Activation=relu & 1,625 & (25)\\
      \hline
      Batch normalization & - & 100& (25)\\
      \hline
      Fully connected & Units=1, Activation=sigmoid & 26& (1)\\
      \hline
    \end{tabular}
  }
  \caption{Our   recurrent  model   hyper-parameters,
    parameters and tensor sizes. Layers in bold with cyan background are sharing weights for both inputs.}
  \label{tab:lstm_model}
\end{table*}

It is important to note that we also  tried to represent our inputs data  using an embedding
layer  instead of  having  a sparse  one-hot  encoding. However,  using
embeddings led  to small  accuracy improvements  only, with  the major
drawback  to greatly  slowing down  the training  process.  Typically,
when our current  recurrent model takes less than 2  hours to train, a
recurrent  model  with embedding  took  more than 3 days to
train,  making   harder  network  architecture   and  hyper-parameters
optimizations.

\section{Results}\label{sec:results}

Before introducing results  of our two models, we  start by explaining
our experimental  methodology.

\subsection{Methodology}

Weights  of each  layer of  our models
have been initialized with Xavier uniform initialization, commonly considered to be a proper way to initialize weights at random values, even if batch normalization decrease the importance of such an initialization.  Both models
have  been  trained  taking  the  binary  cross-entropy  as  the  loss
function, using the Adam optimizer with  a learning rate of 0.001 that
decrease if the validation loss stays  on a plateau for 5 epochs. This
reduction is  done by  multiplying the  learning rate  by 0.9,  and no
reduction is applied  once the learning rate reaches  0.0008. We train
our models  on batches of  2048 samples, large batch  size speeding-up
runtimes and favoring a good batch normalization. We use the front-end
API Keras~\cite{chollet2015keras} with Tensorflow~\cite{tensorflow2015-whitepaper} as a  back-end, and run our experiments on a
NVIDIA GTX 1080 Ti GPU.

\begin{figure*}
  \centering
  \includegraphics[width=0.8\linewidth]{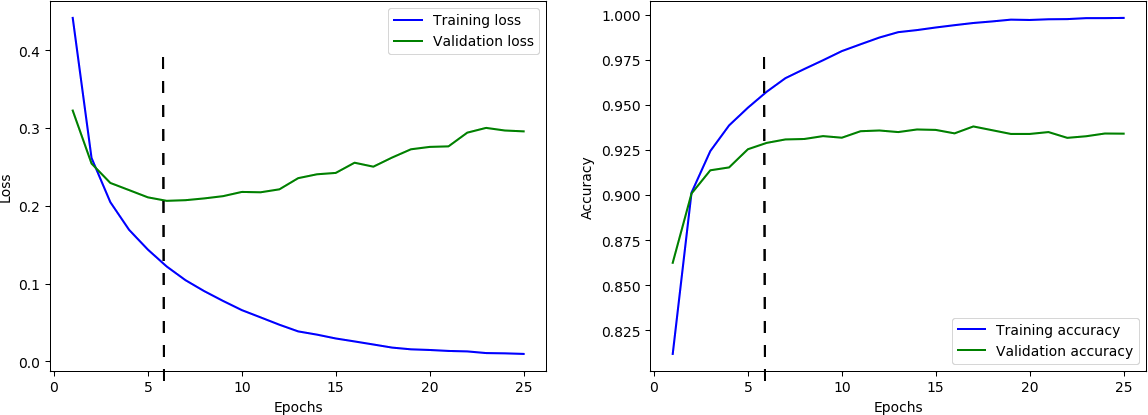}
  \caption{Training and  validation losses  and accuracy of  our fully
    connected model. The dashed  line represents the lowest validation
    loss point,  where trained  weights have been  saved for  our test
    results.}
    \label{fig:fc_curves}
\end{figure*}

\begin{figure*}
  \centering
  \includegraphics[width=0.8\linewidth]{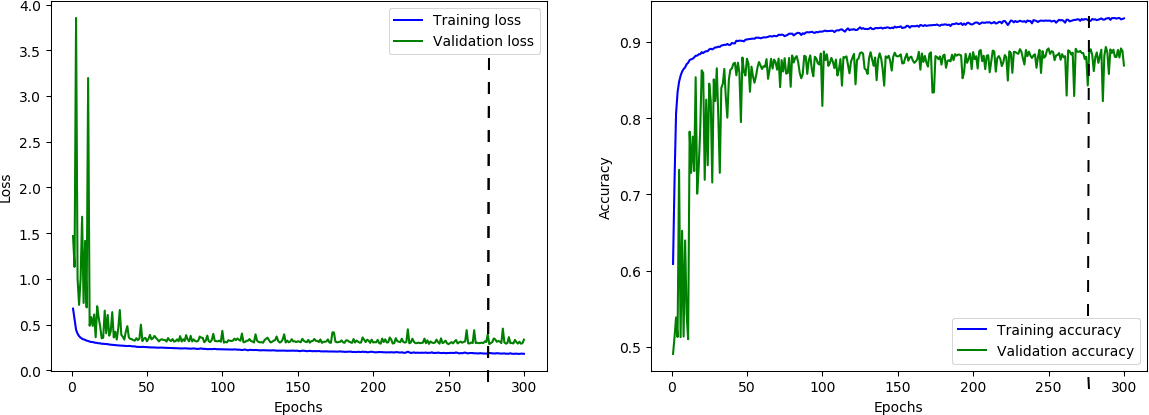}
  \caption{Training  and   validation  losses  and  accuracy   of  our
    recurrent model. The dashed  line represents the lowest validation
    loss point,  where trained  weights have been  saved for  our test
    results.}
  \label{fig:lstm_curves}
\end{figure*}

We  then proceed  as  follows  for both  the  regular  and the  strict
dataset: we  train our  models giving  a training  set and  a hold-out
validation set, and save weights each time the validation loss reaches
a new local  minimum. We then evaluate  our models on the  validation set by
taking weights where  the validation loss was  minimal during training,
represented    by    dash   lines    in    Figures~\ref{fig:fc_curves}
and~\ref{fig:lstm_curves}.  Notice these  weights  do not  necessarily
leads to the best validation accuracy performance, but considering the
moment where validation loss is at its minimum is a good way to avoid having a
model  overfitting  training  data.  Hyper-parameters
optimization   has  been  done   to  improve   validation
accuracy. These  optimizations were conducted by hand,  since no
hyper-parameters optimization tools able  to handle multiple inputs in
Keras exist, up to our knowledge.

Once our hyper-parameters were fixed, we had to consider where the  lowest
validation loss point was reached at epoch number $x$.
We  performed the final tests by retraining
our model  on $x$ epochs taking  both the training and  the validation
sets as our new training set, and testing our models performance on the
hold-out test set. This strict methodology delivers strong robustness to information leaks, that
is, avoiding to indirect overfit the validation set which would bias
our models.   We can  then trust  our final  tests performances  to be
representative of a good generalization behavior of our models.

\subsection{Fully connected model results}

Table~\ref{tab:fc_results} shows results of our fully connected model.
As expected, final  test performances on the  strict dataset (accuracy
of 0.7625,  F-score of 0.5648) are  worst than on the  regular dataset
(accuracy of 0.8997, F-score of 0.9021). But it is more important here
to consider the difference between  the validation performance and the
test performance on  both datasets.  On the  regular dataset, accuracy
and F-score  are similar  for both the validation  and the  test set.
However on the strict dataset, test  performances drop
drastically when compared to validation performances, with an accuracy going
from 0.9289  to 0.7625  and an  F-score from 0.9278  to 0.5648  . This
represents performance drops of 18\% and 39\%, respectively.

These results lead us to  conclude that, even with hold-out validation
and test sets, and even if these sets are perfectly balanced with 50\%
of positive and 50\% of negative samples, the fully connected model is
learning  ``by  heart''  that  some proteins  are  statistically  more
involved into protein interactions.  If such proteins appear both into
the  training and  the test  sets,  the classification  would be  bias
toward  proteins  that  tends  to interact  more,  instead  of  purely
considering extracted features only to perform this classification. It
is then important  to consider test sets with  completely new proteins
for the model, going  far beyond the simple fact to  have test sets of
new  couples of  proteins but  where each  protein appears  separately
somewhere in the training or the validation set. 

\begin{table*}[ht]
  \centering
  {
    \begin{tabular}{|l||c|c||c|c||}
      \cline{2-5}
      \C{} & \multicolumn{2}{c||}{Regular dataset} & \multicolumn{2}{c||}{Strict dataset} \\
      \cline{2-5}
      \C{} & validation set & test set & validation set & test set \\
      \hline
      Accuracy & 0.9073 & 0.8997 & 0.9289 & 0.7625\\
      \hline
      Precision & 0.9284 & 0.9492 & 0.9376 & 0.4269\\
      \hline
      Recall & 0.8867 & 0.8593 & 0.9182 & 0.8345\\
      \hline
      F-score & 0.9071 & 0.9021 & 0.9278 & 0.5648\\
      \hline
    \end{tabular}
  }
  \caption{Results with our fully connected model}
  \label{tab:fc_results}
\end{table*}





\subsection{Recurrent model results}

Table~\ref{tab:lstm_results} shows  results of  our recurrent  model and
confirms  that  this  model  is  more robust  to  overfitting  and  is
generalizing better. The strict dataset being less permissive than the
regular dataset,  we notice  performances drop between  the validation
and the  test set, but  these drops are of  12\% and 26\%  for the
accuracy and the F-score, respectively,  compared to 18\% and 39\% for
the fully connected model.

Moreover,  despite having  worst  accuracy and  F-score  on the  whole
regular  dataset and  on  the  strict validation  set  than the  fully
connected  model,  our  recurrent  model  shows  significantly  better
performances on  the strict test  set, both for accuracy  (0.7833 here
against  0.7625 for  the  first  model) and  for  the F-score  (0.6502
against 0.5648).

We  conclude  that  our  recurrent  model,  with  a  100  times  less
parameters  than our  fully  connected model,  generalizes better  and
then  leads  to  better  predictions.  Its small  size  and  its  good
generalization property  make it  a good candidate  to expand  it with
more layers and/or  more units per layers  if we can train  it on more
data, to improve its global performance.

\begin{table*}[ht]
  \centering
  {
    \begin{tabular}{|l||c|c||c|c||}
      \cline{2-5}
      \C{} & \multicolumn{2}{c||}{Regular dataset} & \multicolumn{2}{c||}{Strict dataset} \\
      \cline{2-5}
      \C{} & validation set & test set & validation set & test set \\
      \hline
      Accuracy & 0.8590 & 0.8623 & 0.8899 & 0.7833\\
      \hline
      Precision & 0.8295 & 0.8791 & 0.8892 & 0.5576\\
      \hline
      Recall & 0.8747 & 0.8443 & 0.8854 & 0.7795\\
      \hline
      F-score & 0.8515 & 0.8614 & 0.8873 & 0.6502\\
      \hline
    \end{tabular}
  }
  \caption{Results with our recurrent model}
  \label{tab:lstm_results}
\end{table*}

\section{Related works}\label{sec:related_work}

In this section, we present recent related works between 2017 and 2019, applying or claiming to apply deep learning methods to prediction protein-protein interactions. We will see that it is not easy to directly compare our results to these papers, either because of a significative methodology difference or because reproducibility of these results is not possible.

Sun et al. in~\cite{Sun2017} are using stack auto-encoders to extract features from protein sequences. The classification predicting protein-protein interaction is then done by directly linking the output of the last auto-encoder to a softmax classifier. For their inputs, there are converting the sequences into fixed-size Boolean vectors, one per sequence, encoding the presence or absence of 3-grams of amino acids, \ie, each possible combination of 3 amino acids. They trained their model doing a 5-fold or 10-fold cross validation, depending their datasets.

Du et al. proposed in~\cite{Du2017} a plain fully connected neural network, similar to our first model in this paper but significantly bigger, with layers containing 512, 256 and 128 units for what should be feature extraction, and 128 units for the head of their network. However, they are not giving as input to the network protein sequences but a list of features, such as  sequence-order descriptors and composition-transition-distribution descriptors, that authors extracted themselves. They used an hold-out validation set together with a test set but then switched for a 5-fold cross validation when comparing their model with others on some other datasets.

Lei et al. use in~\cite{Lei2018} a Deep Polynomial Network on features extracted by hand, like amino acid mutation rates or hydrophobic properties of proteins, to make their classification. Thus, they do not use the  chain of amino acid residues as an input. They based the learning process on a 5-fold cross validation without test sets.

In~\cite{Li2018}, Li et al. present a model composed of an embedding layer, three convolutions and a LSTM layer for feature extractions of protein sequences, before concatenating LSTM output of both proteins and performing classification with a fully connected layer linked to a sigmoid classifier. The architecture of our recurrent model is then near to their model, modulo the embedding and hyper-parameters. It is important to notice they do not mention to apply any regulation method to train their network. Their inputs are also sequence-based and they apply a 5-fold cross validation during training. with a hold-out test set. Interestingly, they also pad to zero their inputs, so they must mask these zeros into the embedding layer to make it learn something (otherwise this layer will interpret zeros as real data and their large number would prevent this layer to learn any good representation of the input). They use convolution layers after the embedding, and like for this paper, the use Keras as an API front-end to program their model. However the current implementation of convolution layers in Keras does not accept zero-masked data, and the authors do not write in their paper how did they manage to get around this technical issue.

Hashemifar et al.~\cite{Hashemifar2018} propose a convolution-based model where feature extractions are terminated by processing data through an original randomly-initialized and untrained matrix they named "random projection module". Their model is also sequence-based but they completely transform inputs to become probabilistic position-specific profiles. To do so, they give peptide sequences to PSI-BLAST to compute these profiles. They are doing 5-fold or 10-fold cross validation, depending of datasets, and have no test sets. We observe that they are the only one tackling the protein-protein interaction prediction by doing a regression instead of a classification. This does not change fundamentally a model or its training, but it allowed them to compute precision-recall curves while focusing the analysis of their results mainly on these curves.

Finally, Zhang et al.~\cite{Zhang2019} present a fully connected model regulated by dropouts. Like~\cite{Du2017}, they use composition-transition-distribution descriptors as features. They apply a 5-fold cross validation and have no separated test sets.

We can also mentioned the work of Zhao et al~\cite{Zhao2018}, proposing a multi-layered LSTM model to predict interface residue pair interactions, thus at a finer level level than prediction interaction between two proteins. This is a direction towards which  we would like to extend our results.

\subsection{Discussions}

With features extracted by hand,~\cite{Du2017, Lei2018} and~\cite{Zhang2019} are missing the point of the specificity and advantage of deep learning over classical machine learning methods: Deep Learning architectures can automatically extract features from raw data and base their training process on them. This allow machines to isolate non-trivial patterns that a human being won't see or is not aware of.

It must be also stressed that only~\cite{Du2017} and~\cite{Hashemifar2018} are giving the source code of their methods (unfortunately,~\cite{Du2017}'s source code is already unavailable) together with their experimental data (partial data for~\cite{Hashemifar2018}). It was then not possible to directly compare methods from papers above on our data following our strict experimental methodology. Although~\cite{Hashemifar2018}'s code is available, their paper does not give enough information about their data transformation through PSI-BLAST to do the same with our data.

None of the papers mentioned above are giving the parameter count of their model, and by judging the size of their model's layers, they must be bigger than our large fully connected model from some orders of magnitude. However, the size of their datasets is almost always smaller than our two datasets (with the exception of a series of generated datasets from~\cite{Hashemifar2018} with a maximum of 500,000 samples, although the exact size is not given). Those models above have then a number of parameters way above the number of training samples, making models prone to overfiting. Only~\cite{Du2017, Sun2017} and~\cite{Li2018} are using hold-out final test sets, but~\cite{Sun2017} seems to give their training accuracy results instead of their test accuracy results, and they change hyper-parameters in function of the used dataset, which makes no sense to evaluate a model (hyper-parameters are part of the model).  \cite{Li2018}'s test sets do contain positive samples only, which is a very bad way to test a model's performance since a model always predicting interactions whatever the protein would have a 100\% accuracy score.

Finally,~\cite{Sun2017} and~\cite{Li2018} are making negative samples by taking two proteins from different subcellular locations. Although this seems to be a good idea at first glance, if negative samples are only of that nature, it may mislead the model to learn how to discriminate if proteins are from the same subcellular locations rather than if they can interact each other.

\section{Conclusion}\label{sec:conclusion}
Our work on protein-protein interactions was performed with a lot of attention for dataset setup and deep learning model architecture to avoid artifacts leading to overfitting on data or to avoid deep learning workflow misuse.
For the two deep learning models presented here, we were able to obtain a rather high degree of predictions with an accuracy of 76.25\% for the fully-connected model and 78.33\% for the recurrent model on the hold-out test set from our {\it strict} dataset, where each couple of proteins is composed of at least one protein that never appears neither in the training set nor in the validation set. These high success rates have been obtained through a more drastic methodology than other methods discussed above, in order to avoid both overfitting and information leak and thus having a model able to scale up on more data in the future. We expect our approach to be useful for the prediction of PPI for many other organisms in the perspective of whole proteome modelling.

An interesting extension of this work would be to focus on the prediction of interface residue pair interactions like in~\cite{Zhao2018}. Being able to predict interactions at such a low level in sequences would bring new perspectives to the understanding of how living cells work.

\end{document}